# A Systematic Study of Retrieval Pipeline Design for Retrieval-Augmented Medical Question Answering

Nusrat Sultana[1,*], Abdullah Muhammad Moosa [1], Kazi Afzalur Rahman [2], Sajal Chandra Banik [2]

[1] Department of Mechatronics & Industrial Engineering, Chittagong University of Engineering & Technology, Chittagong, Bangladesh.

[2] Department of Mechanical Engineering, Chittagong University of Engineering & Technology, Chittagong, Bangladesh.

* Correspondence: nusratsultana@cuet.ac.bd

**Abstract:** Large language models (LLMs) have demonstrated strong capabilities in medical question answering; however, purely parametric models often suffer from knowledge gaps and limited factual grounding. Retrieval-augmented generation (RAG) addresses this limitation by integrating external knowledge retrieval into the reasoning process. Despite increasing interest in RAG-based medical systems, the impact of individual retrieval components on performance remains insufficiently understood. This study presents a systematic evaluation of retrieval-augmented medical question answering using the MedQA USMLE benchmark and a structured textbook-based knowledge corpus. We analyze the interaction between language models, embedding models, retrieval strategies, query reformulation, and cross-encoder reranking within a unified experimental framework comprising forty configurations. Results show that retrieval augmentation significantly improves zero-shot medical question answering performance. The best-performing configuration was dense retrieval with query reformulation and reranking achieved 60.49% accuracy. Domain-specialized language models were also found to better utilize retrieved medical evidence than general-purpose models. The analysis further reveals a clear tradeoff between retrieval effectiveness and computational cost, with simpler dense retrieval configurations providing strong performance while maintaining higher throughput. All experiments were conducted on a single consumer-grade GPU, demonstrating that systematic evaluation of retrieval-augmented medical QA systems can be performed under modest computational resources.

**Keywords:** Retrieval-Augmented Generation; Medical Question Answering; Large Language Model; Information Retrieval Optimization;

# 1. Introduction

Recent advances in large language models (LLMs) have significantly improved performance across a wide range of natural language processing tasks, including question answering, reasoning, and information extraction [1], [2], [3].Transformer-based architectures have demonstrated strong capabilities in biomedical and clinical natural language processing, enabling improved performance on tasks such as medical question answering and clinical reasoning[4], [5], [6]. Benchmarks including MedQA, PubMedQA, and MedMCQA have become widely used for evaluating medical knowledge understanding and reasoning ability in language models[7], [8], [9]. These developments have opened new opportunities for applying LLMs to applications such as clinical decision support, medical education, and biomedical knowledge retrieval.

Despite these advances, purely parametric language models remain limited by their reliance on knowledge encoded during pretraining. Medical knowledge evolves rapidly, and models may generate outdated or incorrect responses when relevant information is not well represented in their training data[10]. In addition, hallucination and lack of factual grounding remain major challenges for deploying language models in safety-critical domains such as healthcare[11], [12], [13]. Retrieval-augmented generation (RAG) has emerged as a promising approach to address these limitations by integrating external knowledge retrieval with language model reasoning[14]. In a RAG framework, relevant documents are retrieved from an external knowledge base and provided to the language model as contextual evidence during answer generation. This approach allows models to ground their responses in explicit knowledge sources rather than relying solely on internal parameters. As a result, RAG has been increasingly applied in biomedical natural language processing tasks such as medical question answering, biomedical literature retrieval, and clinical decision support systems [15], [16], [17], [18].

Recent work has demonstrated that combining large language models with external retrieval mechanisms can improve factual grounding and reasoning performance in medical tasks [19], [20], [21], [22]. Domain-specialized models such as Med-PaLM have shown that integrating retrieval with medical LLMs can significantly improve the quality and reliability of responses to complex medical questions [20]. Nevertheless, many existing studies evaluate only a single retrieval configuration or focus primarily on model architecture improvements.

Several key questions remain open. The interaction between embedding models, retrieval pipelines, reranking strategies, and query reformulation is not well understood in medical QA. Additionally, few studies isolate the contribution of individual retrieval components within a unified framework, and the tradeoff between retrieval effectiveness and computational cost remains underexplored.To address these gaps, we conduct a systematic evaluation of retrieval-augmented medical QA on the MedQA USMLE benchmark using a structured textbook-based corpus. We implement a multi-stage retrieval pipeline with query reformulation, dense or hybrid retrieval, and cross-encoder reranking, and analyze its interaction with both domain-specific and general-purpose LLMs.

Our results show that retrieval augmentation significantly improves zero-shot performance. Dense retrieval with reformulation and reranking achieves the highest accuracy, while simpler configurations offer strong efficiency–performance tradeoffs. We further demonstrate that such evaluations can be conducted on a single consumer-grade GPU, highlighting the practicality of our approach. Overall, this work provides a concise empirical analysis of medical RAG systems and identifies key design considerations for building efficient and effective retrieval pipelines.

In summary, our contributions are five-fold:

• We introduce a unified experimental framework for systematically analyzing retrieval-augmented medical QA systems, enabling controlled evaluation of interactions between language models, embedding models, retrieval strategies, and prompting methods on the MedQA USMLE benchmark.

• We propose a structure-preserving medical corpus construction pipeline that maintains textbook hierarchy and sentence boundaries during chunking, ensuring semantically coherent retrieval units for improved knowledge grounding.

• We develop and analyze a multi-stage retrieval pipeline incorporating query reformulation, dense or hybrid retrieval, and cross-encoder reranking, allowing fine-grained evaluation of individual retrieval components.

• We conduct a comprehensive efficiency-performance tradeoff analysis, identifying retrieval configurations that balance strong accuracy with practical computational cost and runtime efficiency.

• We provide a reproducible and resource-efficient evaluation pipeline, demonstrating that rigorous medical RAG experimentation can be performed on a single consumer-grade GPU, with all implementations publicly available to support future research.

## 2. Related Works

### 2.1. Medical Question Answering with LLMs

Transformer-based language models have significantly advanced biomedical natural language processing tasks [1], [2], [3]. Early domain-specific models such as BioBERT, ClinicalBERT, and PubMedBERT demonstrated that biomedical pretraining substantially improves performance on biomedical text understanding tasks [4], [5], [6]. These models have been widely used for biomedical information extraction, clinical text processing, and medical question answering. Medical question answering benchmarks such as MedQA, PubMedQA, and MedMCQA have played an important role in evaluating the reasoning ability of language models in clinical knowledge domains [7], [8], [9]. Recent instruction-tuned models and prompting techniques such as chain-of-thought reasoning have further improved model performance on complex reasoning tasks [23], [24]. Large-scale medical language models such as Med-PaLM and Med-PaLM 2 have demonstrated strong performance on medical examination benchmarks and clinical reasoning tasks [20]. However, despite these improvements, parametric models remain limited by knowledge cutoff issues and may generate hallucinated responses when reliable grounding information is not available [10], [11], [12], [13].

*2.2. Retrieval-Augmented Generation*

Retrieval-augmented generation combines document retrieval with language model generation to enable models to access external knowledge during inference [14]. Early work demonstrated that integrating retrieval with generative models significantly improves performance on knowledge-intensive NLP tasks [25]. Dense retrieval approaches such as Dense Passage Retrieval (DPR) have become widely used for open-domain question answering systems [28]. Subsequent work has explored hybrid retrieval architectures that combine dense and sparse retrieval signals to improve recall and robustness [26], [27], [28]. More recent studies have also proposed improvements to retrieval pipelines through techniques such as query reformulation, multi-stage retrieval, and cross-encoder reranking [29], [30], [31]. Cross-encoder rerankers refine candidate retrieval results by directly scoring query-document pairs, improving retrieval precision compared to embedding-based similarity measures.

*2.3. RAGs in Biomedical and Medical QA*

Retrieval-augmented approaches have increasingly been applied to biomedical and clinical NLP tasks. Several studies have shown that integrating retrieval mechanisms improves factual grounding and reasoning ability in biomedical question answering systems [15], [16], [17], [18]. Recent work has explored combining large language models with biomedical retrieval systems to improve evidence-based medical reasoning and literature analysis [19], [20], [21], [22]. These systems typically employ dense vector retrieval over biomedical literature or domain-specific knowledge bases. However, many existing studies evaluate only a single retrieval architecture or focus primarily on improving model architectures. As a result, the relative contributions of individual retrieval components such as embedding model selection, query reformulation, reranking strategies, and coarse retrieval filtering remain poorly understood in medical question answering systems.

*2.4. Position of This Work*

In contrast to prior studies that focus on a single retrieval architecture or model configuration, this work performs a systematic evaluation of retrieval-augmented medical question answering pipelines. We analyze the interaction between language models, embedding models, retrieval strategies, and prompting techniques within a unified experimental framework. Specifically, this study evaluates a multi-stage retrieval architecture incorporating query reformulation, dense and hybrid retrieval strategies, and cross-encoder reranking. Using a structured textbook-based knowledge corpus and the MedQA benchmark, we investigate how individual retrieval components influence both accuracy and computational cost. By analyzing multiple retrieval configurations and reporting both performance and efficiency metrics, this work provides a more comprehensive understanding of retrieval pipeline design choices for medical question answering systems.

## 3. Methodology

This study evaluates retrieval-augmented generation (RAG) for medical multiple-choice question answering using the MedQA USMLE benchmark [7], which contains 1273 clinical examination questions. Each question consists of a medical vignette followed by multiple answer options. Model performance was evaluated using exact-match accuracy, while computational efficiency was measured using runtime and throughput. **Figure 1** shows the overall workflow of the proposed retrieval-augmented medical question answering system. A user query is reformulated and encoded using embedding models to retrieve relevant passages from the textbook knowledge base through dense or hybrid retrieval. The retrieved candidates may then be reranked using a cross-encoder, after which the selected evidence is provided to the language model to generate the final answer. To analyze the impact of retrieval augmentation and system design choices, we evaluated multiple combinations of language models, embedding models, retrieval strategies, and prompting methods. In total, 40 experimental configurations were evaluated, enabling a systematic comparison of retrieval components and model behavior.

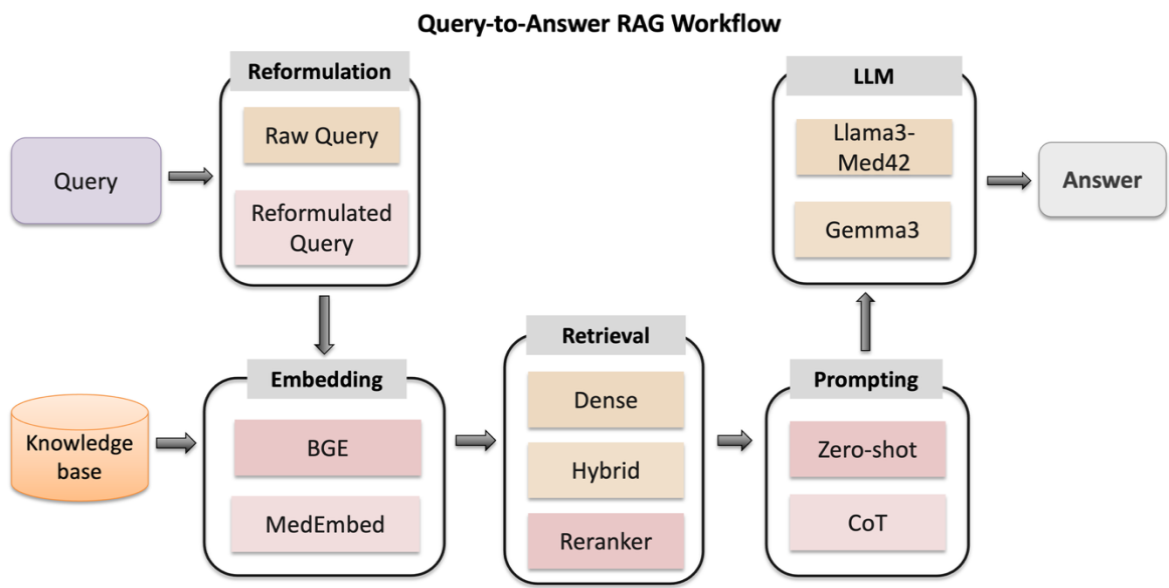

**Figure 1.** End-to-end workflow of the proposed retrieval-augmented medical question answering system.

Two instruction-tuned language models were evaluated: LLaMA3-Med42-8B, a domain-specialized model pretrained on medical corpora and Gemma3, a general-purpose instruction-tuned language model. Both models were evaluated using zero-shot prompting, while LLaMA-Med42 was additionally tested using Chain-of-Thought (CoT) prompting to examine the effect of structured reasoning. However, CoT prompting was not evaluated across all experimental configurations due to computational time constraints, as CoT inference substantially increases runtime compared to zero-shot prompting.

*3.1. Knowledge Corpus and Chunk Construction*

The retrieval knowledge base consisted of curated medical textbooks [7]. To construct retrieval units suitable for semantic search, the corpus was processed using a structure-preserving chunking strategy as shown in **Figure 2**.

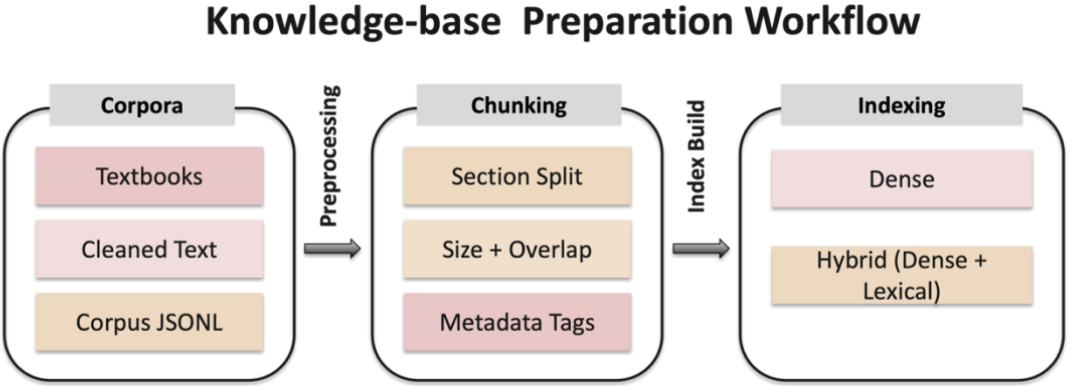

**Figure 2.** Corpora to chunking and indexing workflow.

The preprocessing pipeline is illustrated in **Figure 3**. First, raw medical textbook files were cleaned by removing figures, references, URLs, and formatting artifacts. The text was then structurally segmented into chapters and sections. Within each section, paragraphs were split and filtered to remove extremely short fragments. Finally, chunks were generated within section boundaries using token-window constraints and sentence-aware termination, ensuring that each chunk contained coherent clinical context.

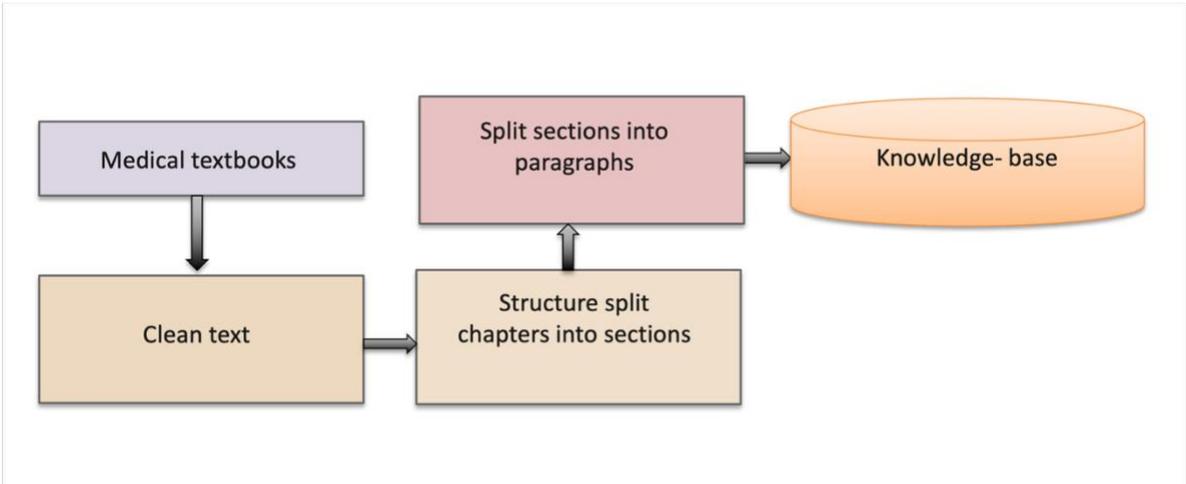

**Figure 3.** Section-wise Chunking Workflow.

Chunks were never allowed to end mid-sentence. If a token limit was reached before a sentence boundary, chunk termination was rolled back to the nearest preceding full stop. This strategy preserves semantic coherence and ensures that each retrieval unit contains complete clinical context. In preliminary experiments, we also evaluated fixed-token chunking without structural constraints, which produced results comparable to the NO RAG baseline. Each chunk was stored as a JSONL record containing both text and metadata, including textbook name, chapter, section, and token count.

*3.2. Index Construction and Multi-Stage Retrieval Pipeline*

Two dense embedding model configurations were used to construct two separate vector indexes: BAAI/bge-large-en-v1.5 and MedEmbed-large-v0.1. Vector indexes were built using FAISS, where cosine similarity was implemented as inner-product search over normalized embeddings. For hybrid retrieval experiments, a BM25 sparse index was also constructed using the *rank_bm25* implementation. In hybrid mode, sparse and dense rankings were combined using Reciprocal Rank Fusion (RRF). All indexes were constructed offline and stored with the corpus to enable deterministic retrieval during evaluation.

During inference, question answering was performed using a three-stage retrieval pipeline:
1. **Coarse section filtering**: A preliminary retrieval step optionally restricts search to the most relevant textbook sections.
2. **Fine-grained chunk retrieval**: Dense vector retrieval identifies candidate chunks from the indexed corpus. In hybrid mode, BM25 and dense scores are combined using reciprocal rank fusion.
3. **Cross-encoder reranking**: Retrieved chunks are re-ranked using a cross-encoder that scores question-chunk pairs.

To balance retrieval recall and contextual relevance, the system first retrieves up to 150 candidate passages from the corpus. These candidates are then reranked using a cross-encoder, and the top six passages are selected as evidence. The selected passages are packed into a maximum context window of 1200 tokens before being provided to the language model. This design improves retrieval recall while limiting excessive context length, as empirical observations indicated that providing substantially larger context windows can reduce answer accuracy due to increased noise in the input evidence.

Additionally, an optional query reformulation module generates retrieval-optimized medical queries using the language model. The reformulated query is used as an additional retrieval query rather than replacing the original question. As illustrated in **Figure 4**, the reformulation prompt guides the model to convert narrative clinical vignettes into concise textbook-style medical queries by removing non-essential patient details and emphasizing clinically relevant concepts and mechanisms. This process produces terminology aligned with medical textbooks, improving the alignment between the query and knowledge-base passages and thereby increasing the accuracy of retrieved evidence in the RAG pipeline.

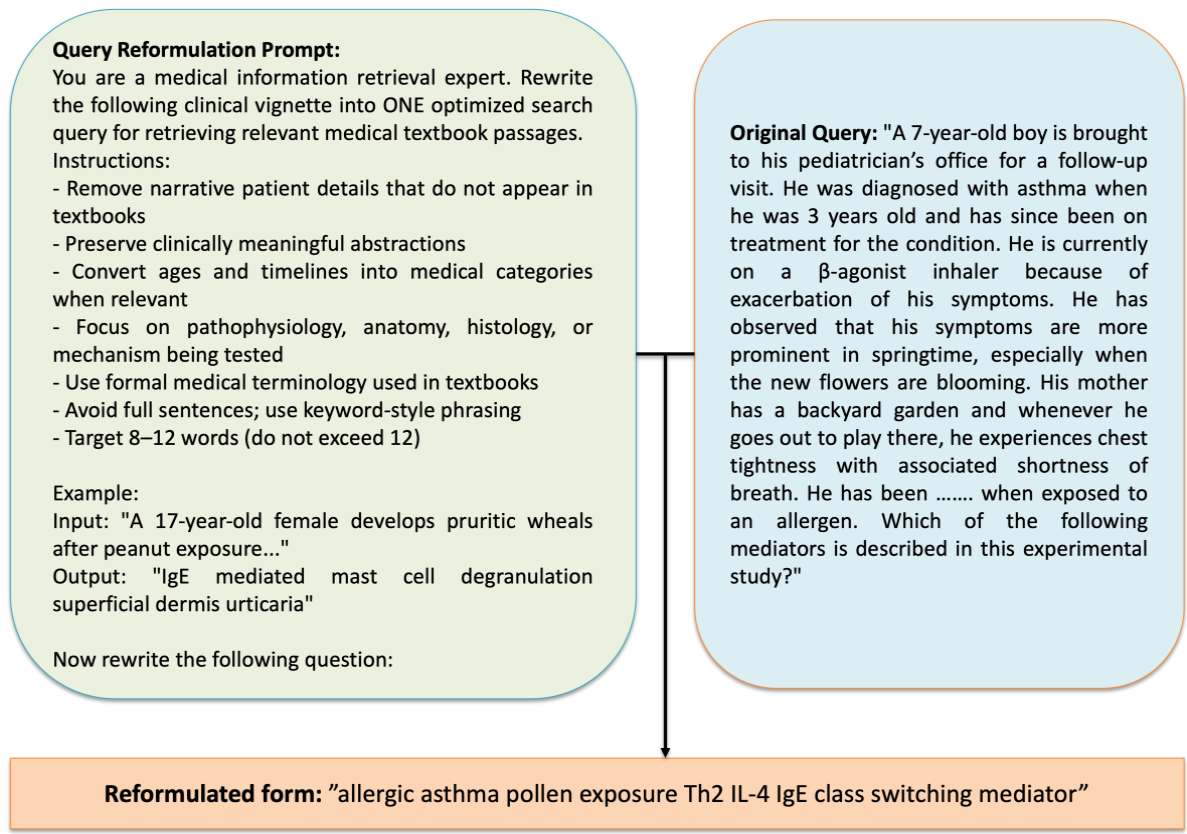

**Figure 4.** Query reformulation example showing how a clinical vignette is converted into a concise textbook-style medical query.

*3.3. Prompting Strategies*

Two prompting approaches were evaluated: Zero-shot prompting, where the model directly predicts the correct answer option and Chain-of-Thought (CoT) prompting, where the model generates intermediate reasoning before selecting the final answer. CoT prompting was evaluated for selected configurations to examine the interaction between explicit reasoning and retrieval augmentation.

*3.4. Experimental Setup and Computational Resources*

All experiments were conducted on a single NVIDIA RTX 3090 GPU (24 GB VRAM) with an Intel Core i9 CPU and 32 GB system RAM. The evaluation framework was designed to operate within modest computational resources while supporting systematic analysis across multiple retrieval configurations.

Despite this constrained hardware setting, the experimental pipeline enabled reproducible evaluation of numerous model and retrieval combinations. Runtime and throughput were recorded for each experiment to analyze the tradeoff between model accuracy and computational efficiency. The complete implementation of the experimental pipeline is publicly available in a GitHub repository to facilitate reproducibility and further research.

## 4. Results

*4.1. Overview*

We evaluated retrieval-augmented medical question answering across multiple experimental axes, including language model choice, embedding model, retrieval configuration, query reformulation, reranking, and prompting strategy. Experiments were conducted on the MedQA USMLE benchmark (n = 1273). Performance was measured using exact-match accuracy, while computational efficiency was evaluated using total runtime and throughput. The full set of evaluated configurations is summarized in **Appendix Table A1**, while this section focuses on representative results and key comparisons.

*4.2. Baseline Performance Without Retrieval*

Zero-shot performance was similar for the two models, with Gemma3 achieving slightly higher accuracy than LLaMA-Med42. Baseline results without retrieval augmentation are summarized in **Table 1**.

Table 1. Baseline Performance Without RAG

| Model | Prompting | Accuracy | Runtime (s) | Throughput (ex/s) |
|---|---|---|---|---|
| Gemma3 | Zero-shot | 0.5593 | 148.0 | 8.603 |
| LLaMA-Med42 | Zero-shot | 0.5554 | 94.9 | 13.420 |
| LLaMA-Med42 | CoT | 0.5970 | 3154.3 | 0.404 |

Applying chain-of-thought (CoT) prompting to LLaMA-Med42 improved accuracy to 59.70%, demonstrating the benefit of structured reasoning prompts. This improvement came at a substantial computational cost, increasing total runtime by more than 30× relative to zero-shot inference.

*4.3. Retrieval-Augmented Performance*

Representative retrieval-augmented configurations are shown in **Table 2**. The best-performing configuration used dense retrieval with both query reformulation and cross-encoder reranking, achieving 60.49% accuracy, which represents an absolute improvement of approximately 4.95 percentage points over the zero-shot LLaMA baseline. The confidence intervals for the reported accuracies are shown in Table 2, and indicate that dense retrieval configurations consistently achieve higher performance than the baseline settings.

**Table 2.** Retrieval-Augmented Performance

| Model | Retrieval | Reformulation | Reranker | Accuracy | C.I. | Runtime(s) |
|---|---|---|---|---|---|---|
| LLaMA-Med42 | Dense RAG | Off | Off | 0.5907 | 56.37% - 61.77% | 387.5 |
| LLaMA-Med42 | Dense RAG | On | Off | 0.6002 | 57.32% - 62.71% | 761.5 |
| LLaMA-Med42 | Dense RAG | On | On | **0.6049** | 57.80% - 63.17% | 843.7 |
| LLaMA-Med42 | Hybrid RAG | On | On | 0.5931 | 56.61% - 62.01% | 3079.2 |
| Gemma3 | Dense RAG | On | On | 0.5860 | 55.90% - 61.31% | 1149.2 |

Although several configurations show partially overlapping confidence intervals, the best-performing dense retrieval setup exhibits a clear upward shift in the interval range. To further assess whether these improvements are statistically significant, McNemar's test was applied to paired predictions.

*4.4 Statistical Significance of Retrieval Improvements*

To determine whether retrieval augmentation significantly improved performance, we applied McNemar's test to paired predictions.

**Table 3.** Statistical Comparison of Retrieval Augmentation

| Comparison | Accuracy Δ | McNemar p-value | Interpretation |
|---|---|---|---|
| LLaMA Zero-shot NO RAG vs Dense RAG | +4.95% | 0.00027 | Significant |
| LLaMA CoT NO RAG vs Dense RAG | +0.63% | 0.684 | Not significant |

**Table 3** presents the statistical comparison between baseline and retrieval-augmented configurations using McNemar's test. For the zero-shot setting, retrieval augmentation produced a statistically significant improvement in accuracy, indicating that the RAG configuration corrected substantially more errors than it introduced relative to the NO RAG baseline. In contrast, the accuracy improvement observed under CoT prompting was not statistically significant. This suggests that while retrieval augmentation substantially benefits zero-shot inference, its additional impact is limited when structured reasoning prompts already provide strong parametric guidance.

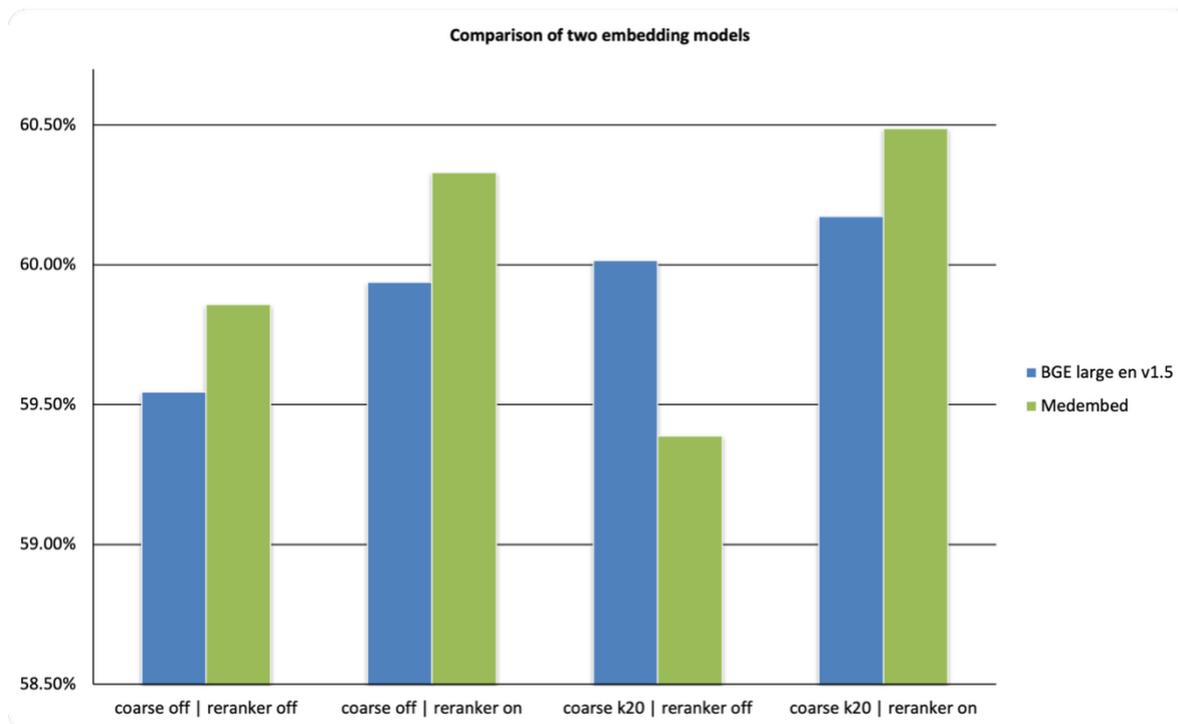

**Figure 5.** Comparison of MedEmbed and BGE under dense retrieval with query reformulation and Llama- Med 42, zero-shot.

**Figure 5** compares the performance of the two embedding models under matched retrieval configurations using dense retrieval with query reformulation and the LLaMA-Med42 model. Overall, MedEmbed outperformed BGE in three of the four configurations, producing a small average improvement of +0.10 percentage points. The largest gain occurred in the coarse off + reranker on configuration ($\Delta$ = +0.0039), indicating that domain-specific medical embeddings can provide modest benefits when combined with reranking. However, the coarse k20 + reranker off configuration slightly favored BGE, suggesting that while medical embeddings can improve retrieval quality, their advantage depends on the overall retrieval pipeline design.

*4.5 Retrieval Component Analysis*

To isolate the contribution of individual retrieval components, we computed paired differences across matched configurations in **Table 4**.

**Table 4.** Technique-Level Accuracy and Runtime Deltas

| Technique | Δ Accuracy | Δ Runtime (s) |
|---|---|---|
| Reranker (On vs Off) | +0.0135 | +107.19 |
| Reformulation (On vs Off) | +0.0091 | +451.33 |
| Hybrid vs Dense | -0.0185 | +3079.69 |
| Coarse k20 vs Off | +0.0003 | -21.27 |

Positive values in Δ Accuracy indicate that the first configuration listed in each comparison achieved higher accuracy than the second, while negative values indicate reduced performance. Similarly, positive values in Δ Runtime indicate increased execution time relative to the second configuration. Enabling the cross-encoder reranker produced consistent accuracy gains with a modest runtime increase. Query reformulation also improved performance but introduced higher computational overhead. In contrast, hybrid retrieval decreased accuracy while substantially increasing runtime, indicating that sparse-dense fusion did not provide benefits for the structured textbook corpus used in this study. Coarse section filtering had negligible impact on both accuracy and runtime.

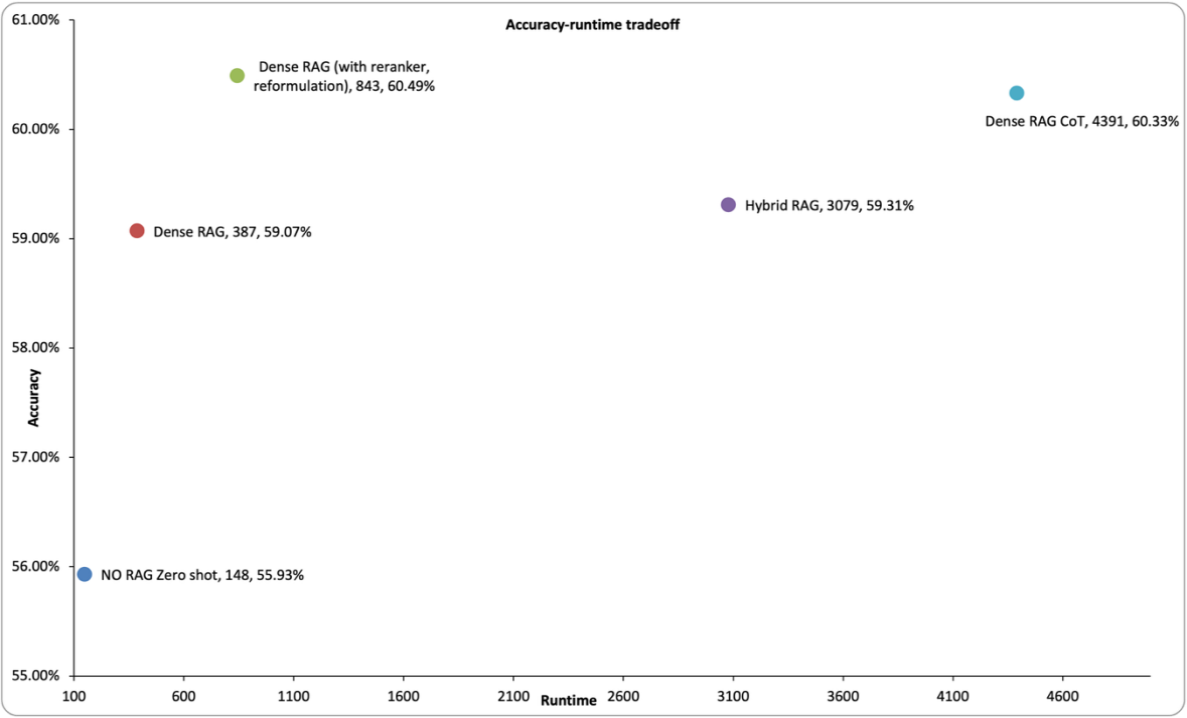

**Figure 6.** Accuracy-runtime tradeoff across retrieval configurations (run time in second, log scale).

**Figure 6** illustrates the tradeoff between accuracy and runtime across the evaluated configurations. Moving from the NO RAG zero-shot baseline to dense RAG results in a notable improvement in accuracy with only a moderate increase in execution time. The dense RAG configuration with query reformulation and reranking achieves the most favorable balance between accuracy and runtime, representing the best overall efficiency-performance tradeoff among the evaluated methods. In contrast, hybrid RAG and dense RAG with CoT prompting incur substantially higher runtime while providing little or no additional accuracy improvement compared to the best dense retrieval configuration.

*4.6 Summary of Findings*

Across all experiments, several consistent patterns emerged:

1. Retrieval augmentation significantly improved zero-shot medical question answering performance.
2. Domain-specialized LLaMA-Med42 achieved higher peak performance than the general-purpose Gemma3 model.
3. Query reformulation and cross-encoder reranking provided the largest retrieval gains.
4. Hybrid sparse–dense retrieval did not outperform dense retrieval in this structured textbook corpus.
5. Coarse section filtering had minimal impact on performance.
6. A practical speed–accuracy tradeoff exists, with simple dense RAG configurations offering strong performance at substantially lower computational cost.

## 5. Discussion

This study provides a systematic evaluation of retrieval-augmented generation for medical question answering using a structured textbook knowledge base. The results demonstrate that retrieval augmentation can substantially improve the performance of large language models on clinical reasoning tasks while maintaining practical computational requirements.

First, retrieval augmentation significantly improved zero-shot medical question answering performance. The best-performing configuration achieved 60.49% accuracy, representing an improvement of approximately five percentage points over the zero-shot baseline. Statistical testing confirmed that this improvement is significant. This finding supports prior work showing that integrating external knowledge retrieval can compensate for the limitations of purely parametric models by grounding model predictions in explicit medical evidence. Second, the results highlight the importance of domain specialization in language models. Across comparable retrieval configurations, the domain-specialized LLaMA-Med42 model consistently achieved higher peak accuracy than the general-purpose Gemma3 model. This suggests that models pretrained on biomedical corpora are better able to interpret and integrate retrieved clinical evidence during reasoning. The results therefore reinforce the value of combining domain-adapted language models with retrieval augmentation for medical applications. Third, the study demonstrates the importance of retrieval pipeline design. Query reformulation and cross-encoder reranking both produced consistent improvements in accuracy. Reformulation improves alignment between narrative clinical questions and the terminology used in medical textbooks, while reranking improves the precision of retrieved evidence by evaluating question-passage pairs more directly. In contrast, hybrid sparse-dense retrieval did not outperform dense retrieval in the evaluated setting. This result likely reflects the structured nature of the textbook corpus, where semantic embeddings are already sufficient to capture relevant conceptual relationships.

Another key observation concerns the tradeoff between retrieval effectiveness and computational cost. The highest-performing configuration combines dense retrieval, query reformulation, and reranking, but introduces additional runtime overhead. However, a simpler dense retrieval configuration without reranking achieved accuracy close to the best configuration while providing substantially higher throughput. These findings suggest that system designers may select retrieval configurations based on deployment constraints, balancing accuracy gains against computational efficiency. The study also provides insights into context construction for RAG systems. Retrieving a larger candidate pool followed by reranking and restricting the final context to a small number of passages produced stronger performance than providing large unfiltered context windows. This observation indicates that careful evidence selection is more important than simply increasing context length, as excessive context can introduce noise that reduces model accuracy. From a practical perspective, this work demonstrates that comprehensive evaluation of retrieval-augmented medical QA systems can be conducted under modest computational resources. All experiments were performed on a single consumer-grade GPU while still enabling systematic comparison across dozens of retrieval configurations. This suggests that researchers and practitioners can explore retrieval-based medical QA systems without requiring large-scale computational infrastructure.

Despite these findings, several limitations should be acknowledged. First, the retrieval corpus consisted of structured medical textbooks, which may differ from other knowledge sources such as clinical guidelines or biomedical literature. Second, the evaluation focused on a multiple-choice medical benchmark, which does not fully capture the complexity of real-world clinical decision support scenarios. Future work could explore retrieval augmentation using larger biomedical corpora, evaluate additional medical reasoning benchmarks, and investigate adaptive retrieval strategies that dynamically adjust evidence selection during inference.

Overall, this study demonstrates that retrieval augmentation provides a practical and effective approach for improving medical question answering systems. By systematically analyzing retrieval components and computational tradeoffs, this work provides insights that can guide the design of efficient and reliable retrieval-augmented medical AI systems.

## 6. Conclusion

This work investigated retrieval-augmented generation for medical question answering using a structured textbook knowledge base and the MedQA USMLE benchmark. Through a systematic evaluation of multiple retrieval pipelines, embedding models, and prompting strategies, the study demonstrates that retrieval augmentation can significantly improve the performance of large language models on medical examination questions.

The results show that dense retrieval combined with query reformulation and cross-encoder reranking provides the most effective configuration for knowledge grounding in the evaluated corpus. In addition, domain-specialized language models were found to better utilize retrieved medical evidence than general-purpose models. The experiments further highlight the importance of balancing retrieval effectiveness with computational efficiency when designing practical RAG systems. By providing a reproducible experimental framework and detailed analysis of retrieval components, this work contributes insights for the design of efficient retrieval-augmented medical QA systems. Future research may extend this framework to larger biomedical corpora, additional clinical reasoning benchmarks, and adaptive retrieval mechanisms for improved evidence selection.

**Declaration of generative AI and AI-assisted technologies in the manuscript preparation process**

During the preparation of this work, the author(s) used ChatGPT (OpenAI) to assist in improving the clarity, structure, and language of the manuscript. The generated content was carefully reviewed, validated, and edited by the author(s) to ensure accuracy and alignment with the research objectives. The author(s) take full responsibility for the content of the published article.

**Availabilty of Code and Results**

The code, experimental configurations, and generated outputs supporting the findings of this study are publicly available at the following repository, enabling full reproducibility of the proposed pipeline.
GitHub Repository: https://github.com/abdullahmoosa/medrag-research

**Funding Information**

This research received no external funding.

# Appendix

**Table A1: Accuracy and runtime of all evaluated combinations.**

| family | index | retrieval_mode | coarse_mode | reranker | Reformulation | prompt_mode | llm_model | accuracy | runtime |
|---|---|---|---|---|---|---|---|---|---|
| RAG | medembed | dense | on | on | on | Zero shot | llama3 | 0.604870 | 843.7 |
| RAG | bge | dense | on | on | on | cot | llama3 | 0.603299 | 4391.9 |
| RAG | medembed | dense | off | on | on | Zero shot | llama3 | 0.603299 | 847.1 |
| RAG | bge | dense | on | on | on | Zero shot | llama3 | 0.601728 | 844.5 |
| RAG | bge | dense | on | off | on | Zero shot | llama3 | 0.600157 | 761.5 |
| RAG | bge | dense | off | on | on | Zero shot | llama3 | 0.599372 | 834.9 |
| RAG | medembed | dense | off | off | on | Zero shot | llama3 | 0.598586 | 753.0 |
| NO RAG | none | none | none | none | none | cot | llama3 | 0.597015 | 3154.3 |
| RAG | bge | dense | off | off | on | Zero shot | llama3 | 0.595444 | 748.9 |
| RAG | medembed | dense | on | off | on | Zero shot | llama3 | 0.593873 | 758.7 |
| RAG | bge | hybrid | on | on | on | Zero shot | llama3 | 0.593087 | 3079.2 |
| RAG | bge | hybrid | off | on | on | Zero shot | llama3 | 0.591516 | 3074.0 |
| RAG | bge | dense | on | off | off | Zero shot | llama3 | 0.590731 | 387.5 |
| RAG | bge | dense | off | off | off | Zero shot | llama3 | 0.590731 | 377.9 |
| RAG | bge | dense | on | on | off | Zero shot | llama3 | 0.589159 | 467.0 |
| RAG | bge | dense | off | on | off | Zero shot | llama3 | 0.589159 | 466.9 |
| RAG | bge | dense | on | on | off | Zero shot | gemma3 | 0.588374 | 666.2 |
| RAG | bge | dense | off | on | off | Zero shot | gemma3 | 0.588374 | 678.5 |
| RAG | bge | dense | off | on | on | Zero shot | gemma3 | 0.586803 | 1197.5 |
| RAG | bge | dense | on | on | on | Zero shot | gemma3 | 0.586017 | 1149.2 |
| RAG | medembed | dense | on | on | off | Zero shot | llama3 | 0.585232 | 475.3 |
| RAG | medembed | hybrid | on | on | on | Zero shot | llama3 | 0.585232 | 6552.6 |
| RAG | medembed | dense | off | on | off | Zero shot | llama3 | 0.584446 | 476.2 |
| RAG | medembed | dense | off | off | off | Zero shot | llama3 | 0.580518 | 390.2 |
| RAG | medembed | dense | on | off | off | Zero shot | llama3 | 0.578947 | 383.4 |
| RAG | medembed | dense | on | on | on | Zero shot | gemma3 | 0.571877 | 1188.2 |
| RAG | bge | dense | off | off | off | cot | llama3 | 0.571092 | 4027.1 |
| RAG | medembed | dense | off | on | on | Zero shot | gemma3 | 0.568735 | 1475.1 |
| RAG | medembed | dense | on | on | off | Zero shot | gemma3 | 0.567164 | 693.5 |
| RAG | medembed | dense | off | on | off | Zero shot | gemma3 | 0.567164 | 690.8 |
| RAG | bge | dense | off | off | on | Zero shot | gemma3 | 0.564808 | 1125.1 |
| RAG | bge | dense | on | off | on | Zero shot | gemma3 | 0.564022 | 1057.8 |
| RAG | bge | dense | on | off | off | Zero shot | gemma3 | 0.563236 | 575.4 |
| RAG | bge | dense | off | off | off | Zero shot | gemma3 | 0.563236 | 574.5 |
| NO RAG | none | none | none | none | none | Zero shot | gemma3 | 0.559309 | 148.0 |
| NO RAG | none | none | none | none | none | Zero shot | llama3 | 0.555381 | 94.9 |
| RAG | medembed | dense | off | off | on | Zero shot | gemma3 | 0.553810 | 1070.2 |
| RAG | bge | hybrid | off | off | off | Zero shot | llama3_ | 0.553024 | 2514.0 |
| RAG | medembed | dense | on | off | on | Zero shot | gemma3 | 0.553024 | 1092.5 |
| RAG | medembed | dense | on | off | off | Zero shot | gemma3 | 0.537313 | 609.5 |
| RAG | medembed | dense | off | off | off | Zero shot | gemma3 | 0.537313 | 613.8 |